\pdfoutput=1
\documentclass[11pt]{article}

\usepackage[final]{acl}

\usepackage{times}
\usepackage{latexsym}
\usepackage{subfigure}
\usepackage[T1]{fontenc}
\usepackage[utf8]{inputenc}

\usepackage{microtype}
\usepackage{amsmath}
\usepackage{amsthm}
\usepackage{amssymb}
\usepackage{algorithm}

\usepackage{algorithmicx}
\usepackage{algpseudocode}
\usepackage{multirow}
\usepackage{multicol}
\usepackage[switch]{lineno}
\usepackage{inconsolata}

\usepackage{graphicx}

\title{AhaKV: Adaptive Holistic Attention-Driven KV Cache Eviction for Efficient Inference of Large Language Models}

\author{
 \textbf{Yifeng Gu\textsuperscript{1}}, \textbf{Zicong Jiang\textsuperscript{1}}, \textbf{Jianxiu Jin\textsuperscript{1}},  \textbf{Kailing Guo\textsuperscript{1}}, \textbf{Ziyang Zhang\textsuperscript{1}}, \textbf{Xiangmin Xu\textsuperscript{1,2}},
\\
 \textsuperscript{1}South China University of Technology, \textsuperscript{2}Pazhou Laboratory
\\
 \small{
   \textbf{Correspondence:} \href{mailto:email@domain}{guokl@scut.edu.cn}
 }
}

\begin{document}
\maketitle
\begin{abstract}
Large Language Models (LLMs) have significantly advanced the field of Artificial Intelligence. However, their deployment is resource-intensive, not only due to the large number of model parameters but also because the (Key-Value) KV cache consumes a lot of memory during inference. 
While several works propose reducing the KV cache by evicting the unnecessary tokens, these approaches rely on accumulated attention score as eviction score to quantify the importance of the token. We identify the accumulated attention score is biased and it decreases with the position of the tokens in the mathematical expectation. As a result, the retained tokens concentrate on the initial positions, limiting model's access to global contextual information. 
To address this issue, we propose Adaptive holistic attention KV (AhaKV), it addresses the bias of the accumulated attention score by adaptively tuning the scale of softmax according the expectation of information entropy of attention scores. To make use of the holistic attention information in self-attention mechanism, AhaKV utilize the information of value vectors, which is overlooked in previous works, to refine the adaptive score. We show theoretically that our method is well suited for bias reduction. We deployed AhaKV on different models with a fixed cache budget. Experiments show that AhaKV successfully mitigates bias and retains crucial tokens across global context and achieve state-of-the-art results against other related work on several benchmark tasks. 
\end{abstract}

\section{Introduction}
Transformer~\cite{vaswani2017attention} has demonstrated remarkable success in a wide range of domains, including language modeling \cite{kenton2019bert,raffel2020exploring}, image recognition~\cite{dosovitskiy2020image,xie2021segformer}, and speech recognition~\cite{dong2018speech,nakatani2019improving}, owing to its powerful modeling capabilities. In particular, in the field of natural language processing, Transformer-base large language models (LLMs)~\cite{achiam2023gpt,zhang2022opt,touvron2023llama,touvron2023llama2} have become defacto method. With the rise of multi-turn conversations and long document processing scenarios~\cite{bai2023longbench,chen2023longlora}, the lengths of model's input text \cite{peng2023yarn,liu2024world} continue to grow, leading to a significant increase in the deployment costs associated with large-scale models. These costs arise not only from the computational resources required for parameter loading and attention computation, but also from the memory consumption caused by the widely used Key-Value (KV) cache strategy during the generation process~\cite{pope2023efficiently}. The KV cache which stores precomputed keys and values, which is designed avoid re-computation and improve inference speed. However, this comes at the cost of considerable memory overhead, making it a critical bottleneck in efficient model deployment. For example, in LLaMA-2-7B \cite{touvron2023llama2}, when the input batch size is 8 and the context length reaches 32K tokens, the KV cache alone can grow to 128GB, which far exceeds the memory required to store the model parameters themselves.

\begin{figure}[tp]
\centering
\includegraphics[width=8cm]{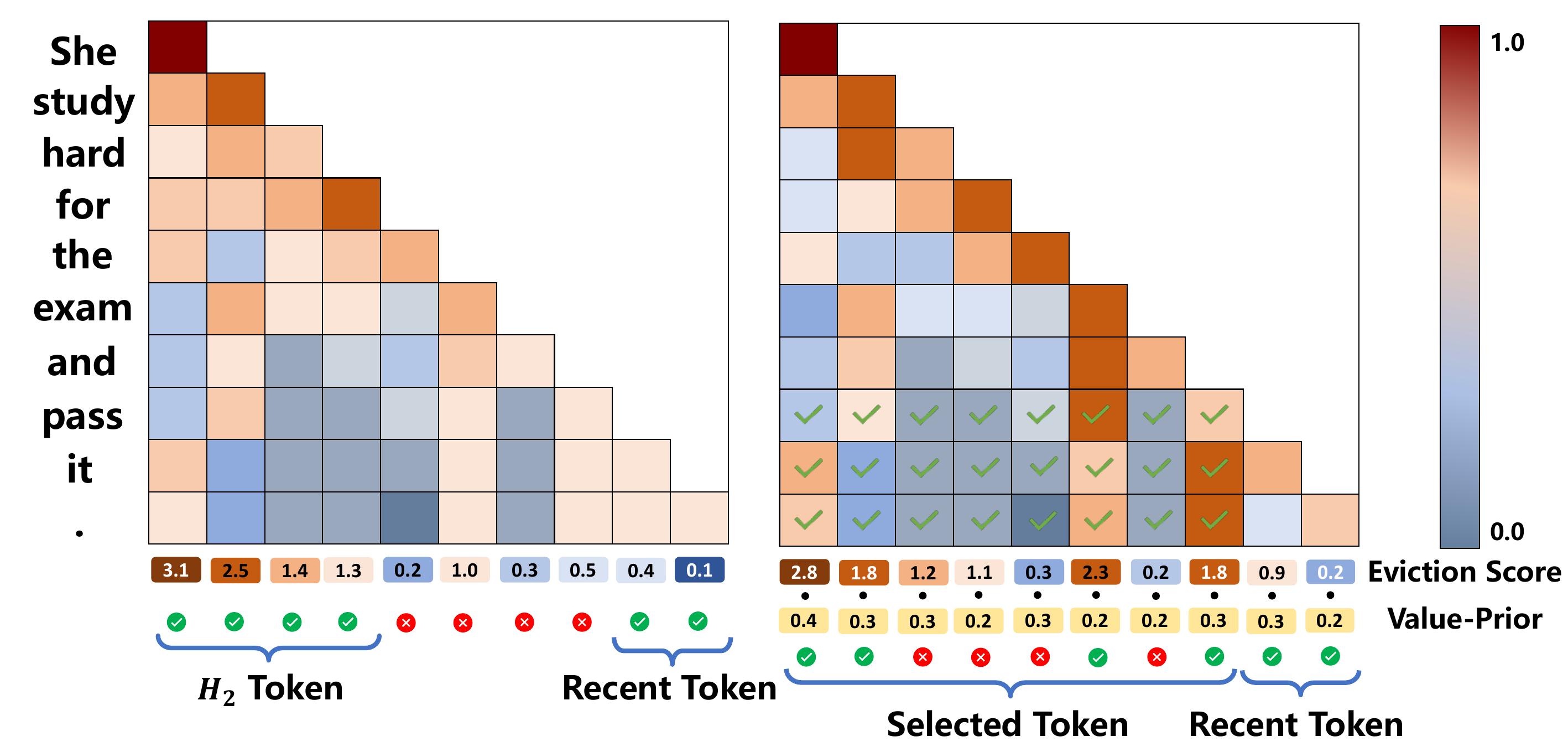}
\caption{Left: The $H_2O$-base method, which uses the accumulated attention score as an eviction score, has a higher eviction score for the token on the left side of the sequence. Right: The AhaKV eviction strategy, which focuses the attention scores on key tokens and selects the attention scores of key tokens to calculate the eviction scores, avoiding positional bias.} \label{Fig.1}
\end {figure}

The substantial KV cache consumption poses significant challenges for deployment on typical consumer GPUs. Recent researches \cite{xiao2023efficient,zhang2024h2o,li2024snapkv,adnan2024keyformer} have shown that retaining only a small subset of crucial tokens' keys and values during generation can achieve performance comparable to those of the full cache LLMs. StreamingLLM \cite{xiao2023efficient} identified that initial tokens play a critical role in maintaining model performance. By retaining only the initial tokens and the most recent tokens, StreamingLLM significantly reduces the KV cache size with promising results. To avoid losing vital information within the middle of the input, $H_2O$~\cite{zhang2024h2o} leveraged accumulated attention scores as the eviction score during generation, enabling the dynamic removal of less important key-value pairs. This flexible and highly successful method has served as the foundation for many research improvements, leading to the development of new approaches. However, in long context tasks, the cumulative attention score is subject to decay bias. SnapKV~\cite{li2024snapkv} uses the recent window's attention score to retain important tokens to improve long context processing. NACL~\cite{chen2024nacl} introduces stochastic eviction to reduce the eviction score bias to improve long context performance. However, these methods do not consider the flattening of attention scores in long contexts, and they do not make full use of numerical information.

However, the accumulated attention scores used in existing methods be further optimized. Figure ~\ref{Fig.1} illustrates the attention scores and accumulated attention scores for a segment of a sentence, demonstrating that the $H_2O$-based approach tends to retain tokens from the earlier part of the sequence. Even when tokens in the middle are semantically crucial, their importance is underestimated within framework. The bias stems from the widely employed causal mask in Transformer-based architectures, which is designed to prevent information leakage from future tokens. Specifically, the causal mask zeros out all values after the current token in the softmax input, ensuring that each token can only attend to itself and preceding tokens. As the sequence length increases, the average attention score for each token decreases after the softmax operation. As depicted in Figure \ref{Fig.1}, tokens on the right side of the sequence are computed fewer times than those on the left. Furthermore, since tokens on the right inherently receive lower attention scores compared to those on the left during the initial computation, their accumulated attention scores are further diminished in expectation. This results in a systematic bias where tokens on the right side are more likely to be evicted, regardless of their actual importance. Second, the accumulated attention score framework only uses information from the queries and keys, neglecting the information about the values in the self-attention mechanism. In fact, values carry the contextual information prediction. Effectively leveraging the information contained in the values is therefore another key consideration when designing an eviction strategy. By incorporating value-based insights, it may be possible to develop a more balanced and context-aware eviction approach that better preserves semantically important tokens across the entire sequence. 

In this paper, we propose Adaptive holistic attention KV (\textbf{AhaKV}), which addresses the bias of the accumulated attention score by adaptively tuning the scale of softmax and makes use of the holistic attention information to evict redundant tokens, reducing memory usage while preserving text processing capacity. We analyze the reason of the bias through theoretical derivation, highlighting that a consistent number of cumulative entries is maintained for each token when computing the eviction score. To counteract the flattening of the attention score as the number of tokens increases, we propose step gain softmax (SG-softmax) to adaptively adjust the attention distribution to emphasize key tokens. Additionally, we compute a prior weight for each token using the modulus length of value to enhance the eviction score. The eviction score in AhaKV avoids positional bias and makes full use of the information of queries, keys, and values (QKV) to maintain the text processing capacity after eviction of the token, and achieves state-of-the-art in comparison to other eviction methods. Overall, Our main contributions are as follows:

\begin{itemize}
\item[(1)] Analyses of self-attention mechanisms explain why cumulative attention scores retain early tokens and provide a theoretical basis for addressing bias. Meanwhile, we propose Recent Accumulation and SG-softmax to avoid bias. 
\item[(2)] Our work pioneers the effective use of value matrix information through a simple transformation that enhances the eviction score using the value prior. This provides a possible idea for future improvements.
\end{itemize}

\section{Related Work}
\label{sec2}
\subsection{Efficient LLMs Inference}
LLMs have a large inference computational cost. This problem has inspired a lot of research in the inference optimization for LLMs, such as the methods \cite{frantar2023sparsegpt,ma2023llm,sun2023simple} adopting pruning to reduce the redundant parameters of LLMs and thus improve the inference speed. In addition to this, many research \cite{frantar-gptq,dettmers2022gpt3,liu2023llm} have used quantization for LLMs to reduce the full-precision LLM parameters to those of a few bits, thus speeding up the inference of the model. In contrast to the above approaches of modifying model parameters, some other works \cite{zaheer2020big,wang2021spatten,zhang2024h2o,hooper2024kvquant,liu2024kivi,dong2024get} start from the bottleneck of the attention module of secondary computational complexity in inference and speed up the inference process by sparse attention. The approaches of these works are orthogonal and can be integrated together. The work studied in this paper, on the other hand, is closely related to sparse attention and focuses on reducing the memory bottleneck, KV cache, that arises in the inference process.

\subsection{Sparse Attention} 
Sparse attention mainly targets the prefill phase of the transformer and omits certain attention operations to improve computational efficiency. Big bird \cite{zaheer2020big} combines random, local and global attention to sparse attention while maintaining accuracy of generation. Longformer \cite{beltagy2020longformer} introduces sliding windows and task-driven global attention to reduce attention computation. Spatten \cite{wang2021spatten} evaluates the importance of each token by calculating the cumulative attention to dynamically prune the token with minimal attention. However, these sparse-attention approaches do not focus on the growing memory problem despite reducing self-attention computation. The KV cache generated by LLMs inference has become a memory bottleneck for inference.

\begin{figure*}[htbp]
\centering
\subfigure[$H_2O$]{%
    \resizebox*{3.6cm}{!}{\includegraphics{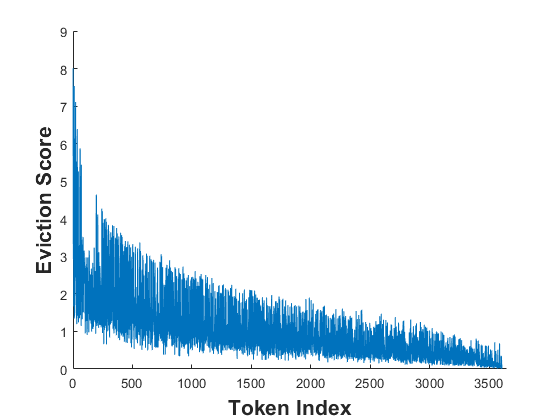}}} \label{fig.2.a}
\subfigure[AhaKV]{%
    \resizebox*{3.6cm}{!}{\includegraphics{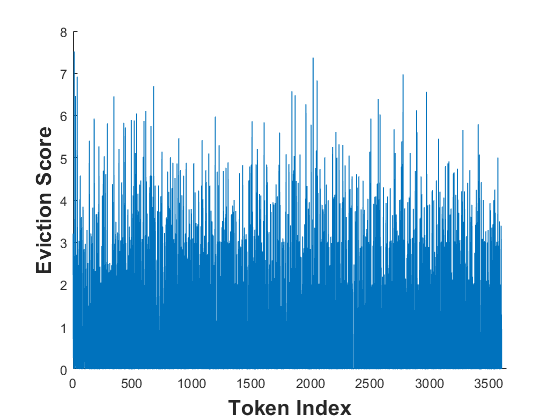}}} \label{fig.2.b}
\subfigure[Retain token index.]{%
    \resizebox*{3.6cm}{!}{\includegraphics{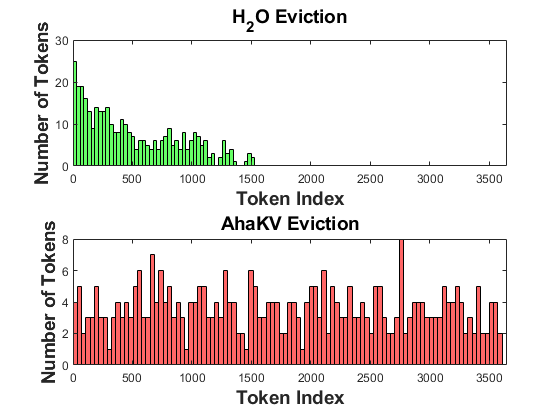}}} \label{fig.2.c}
\subfigure[Sparsity Comparision]{%
    \resizebox*{3.6cm}{!}{\includegraphics{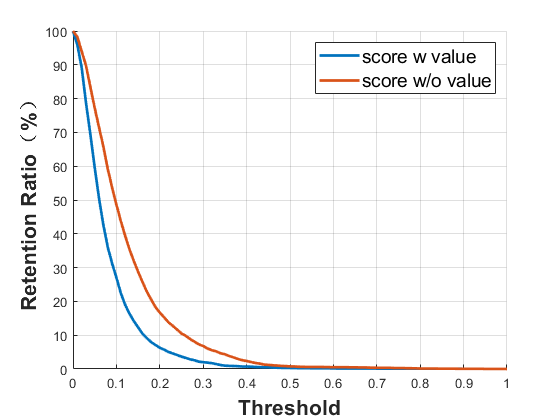}} \label{fig.2.d}} 
\caption{(a,b) shows the eviction scores in different strategy. (c) shows the retention token index. (d) shows the Comparision between Rretention Ratio in different threshold with and without value-prior}
\label{Fig.2}
\end {figure*}

\subsection{KV Cache Eviction} 
Research on KV cache optimization has focused on compressing the out of memory that occurs as the increasingly length of the prompt. In this paper, we focus on KV cache eviction \cite{xiao2023efficient,zhang2024h2o,chen2024nacl,adnan2024keyformer} which focuses on attention sparsity, and explore how to reduce redundantly labelled KVs to reduce memory usage and improve inference efficiency. Which can be subdivided into static eviction and dynamic eviction according to the eviction method. The static eviction is represented by StreamingLLM \cite{xiao2023efficient}, which observes that the beginning token is very important for the whole generation task, and therefore retains the KVs of the 4 initial tokens and the most recent token, and evicts the KVs of the remaining tokens. The study of dynamic eviction focuses on identifying the key tokens and constantly updating the KV cache. The $H_2O$ \cite{zhang2024h2o} method utilises the sum of cumulative attentions as the eviction score and retains the token with the highest score and the nearest token. SnapKV~\cite{li2024snapkv} proposed to compress the long context prompt before calculating the expulsion score to improve the comprehension of long text. NACL \cite{chen2024nacl} introduces random evictions thereby avoid bias in eviction scores. Distinguishing from the above methods, AhaKV avoids the bias of the Eviction Score and makes full use of value to refine the eviction score.

\section{Motivation}

For Decoder-only Large Language Models, the inference process consists of two phases, the prefill phase and the generation phase. In the prefill phase, input tokens are encoded as Key, Value, and Query vectors to compute the self-attention. The Key and Value vectors are typically cached for reusage to accelerate inference. In the generation phase, the model computes attention weights by multiplying the current token’s Query with all cached Key vectors, then applies these weights to the corresponding Value vectors.
With KV caching, only the Key and Value of the new token are computed at each step, while previous vectors are retrieved from the cache. While this approach reduces computational redundancy, it incurs significant memory overhead.

Attention computations exhibit inherent sparsity in practice. Selective eviction of non-critical Key-Value pairs from the cache can reduce memory usage while preserving generation accuracy  \cite{zhang2024h2o,liu2024scissorhands,adnan2024keyformer}. Current eviction strategies predominantly rely on accumulated attention scores \cite{zhang2024h2o}. However, we identify two critical limitations in these approaches.

\subsection{Positional Bias in Accumulated Attention Scores}
\label{sec:3.1}
Figure \ref{Fig.2} analyzes the eviction scores generated by H$_2$O \cite{zhang2024h2o} for an attention head of a layer in the LLaMA2 model, along with the distribution of retained token indices across a 3,600-token sequence. The results reveal an intrinsic positional bias: accumulated attention score has a intrinsic bias, that they gradually decrease with token position. Tokens beyond position 1,500 are disproportionately evicted,
leading to the catastrophic loss of nearly half of the context information. This phenomenon is not unique to the LLaMA model. In fact, it is prevalent across various models, and supporting results from other models are provided in Appendix \ref{appx:C}. It highlights the eviction strategies need to ensure more uniform information retention across all positions.

\subsection{Neglecting Values' Effect}
While Value vectors fundamentally shape self-attention outputs, existing eviction strategies ignore their contribution. We propose augmenting accumulated attention scores with Value-derived prior weights to refine eviction scores generated by H$_2$O. Figure \ref{fig.2.d} compares normalized retention rates for thresholds applied to baseline scores versus  value-enhanced scores. Incorporating value priors obtains lower retained token proportion at equivalent thresholds. This demonstrates that value magnitudes provide complementary signals for identifying semantically important tokens.
\section{Methodology}

In this section, we analyze the limitations of accumulated attention scores and the computation of softmax mathematically. We identify the reason of the bias and mitigate it through theoretical deduction. 
We further refine the eviction score using the value-based prior to make use of the holistic attention. A novel method Adaptive holistic attention KV (AhaKV) that evicts tokens based on unbiased eviction scores while preserving global context information is proposed.

\subsection{Analysis of the Accumulated Attention Score}

Denote the query matrix as $Q \in \mathbb{R}^{n \times d}$, the key matrix as $K \in \mathbb{R}^{n \times d}$, and the hidden dimension as $d$. The attention score between the $i^{th}$ query and the $j^{th}$ key $a_{i,j}$ is computed as following:
\begin{equation}
    a_{i,j} = \text{softmax}(Q_i^{T} K_j/\sqrt{d})=\frac{e^{Q_i^{T} K_j/\sqrt{d}}}{\sum_{j=1}^{i}e^{Q_i^{T} K_j/\sqrt{d}}}  \label{equa:ascore}
\end{equation}

The eviction score $S_j$ in H$_2$O for the  $j^{th}$ token is defined as:
\begin{equation}
    S_j = \sum_{i=1}^{n}a_{i,j} \label{equa:evic}
\end{equation}

Due to the casual mask, $a_{i,j}=0$ when $i<j$. Thus, Eq. (\ref{equa:evic}) can be rewritten as:
\begin{equation}
    S_j = \sum_{i=j}^{n}a_{i,j} \label{equa:evic1}
\end{equation}

The number of cumulated term decreases as  $j$ increases, leading to a monotonic decrease trend in the eviction score, which will be proven in the following. Following the previous works \cite{vaswani2017attention,chi-etal-2023-latent,kazemnejad2024impact}, we assume that the components of $Q_i$ and $K_j$ are independent random variables with the mean as 0 and the variance as 1. Under this assumption,  $a_{i,j}$ has the same expectation and variance for the same row of the attention matrix. We prove the monotonicity of the eviction score as follows: 
{
\begin{align}
    & E(S_{j+1}-S_j)=E(S_{j+1}) - E(S_j) \nonumber \\
    = & E(\sum_{i=j+1}^{n}a_{i,j+1})-E(\sum_{i=j}^{n}a_{i,j})
    \nonumber \\
    =& \sum_{i=j+1}^{n}E(a_{i,j+1})-\sum_{i=j}^{n}E(a_{i,j})  \nonumber\\
    =& \sum_{i=j+1}^{n}[E(a_{i,j+1})-E(a_{i,j})] - E(a_{j,j})
    \label{equa:proof4}
\end{align}
}

Since each element in row $i$ has the same mathematical expectation, the first term of Eq. (\ref{equa:proof4}) is 0. However, the expectation of $a_{j,j}$ is a positive value. Thus, $E(S_{j+1}-S_j)<0$, which means that the eviction score decreases monotonically in expectation. This is validated empirically in Figure \ref{Fig.2}, where the accumulated attention score has an clear positional bias.  

To address this bias, we propose summing the attention scores over the nearest $r$ rows. 
This ensures that the eviction score is not influenced by the number of accumulated terms. The modified eviction score is computed as:
\begin{equation}
    S_j = \sum_{i=n-r}^{n}a_{i,j}
    \label{equa:proof6}
\end{equation}

\subsection{Rethink the Softmax Computation}

\begin{figure*}[ht]
\centering
\includegraphics[width=16cm]{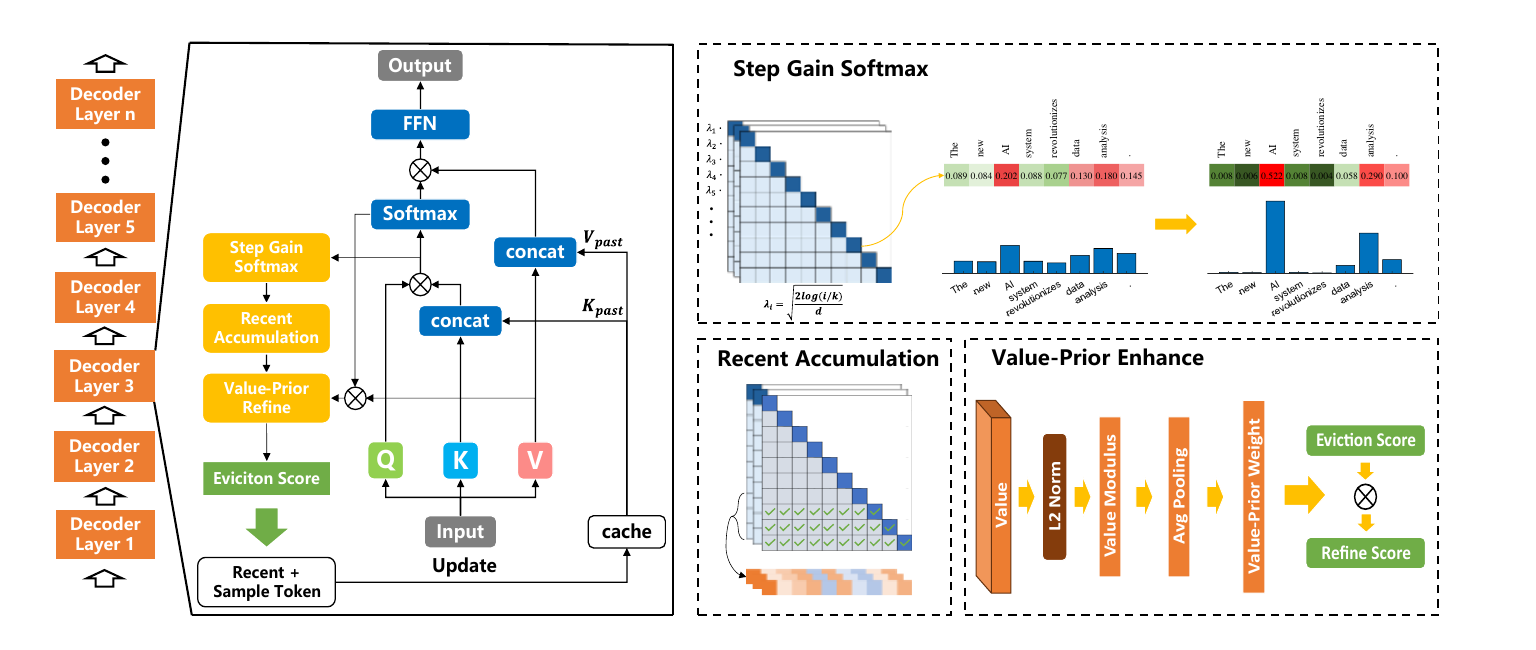}
\caption{AhaKV consists of three main components: SG-softmax, Recent Accumulation, and Value-Prior Refine. First, SG-softmax replaces the standard softmax to prevent attention score flattening. Then, cumulative eviction scores are computed using attention scores from recent rows to ensure each token has the same number of cumulative terms. Next, value-prior is used to refine eviction score selection, and pooling is applied to smooth scores and prevent excessive text sparsification. Finally, tokens are retained based on their eviction scores.}\label{Fig.3}
\end {figure*}

We further analyze the decrease in attention scores as the sequence length grows. As more tokens are included in the softmax computation, the attention scores become diluted, particularly for tokens in later positions. This phenomenon arises because the softmax normalization distributes a fixed total probability mass (summing to 1) across a growing number of tokens, leading to smaller average attention scores for each token.
To better understand the distribution of attention scores, we introduce information entropy from information theory as a complementary measure. Let $H_i$ be the information entropy of the $i^{th}$ row of the attention score matrix, which is given by
\begin{equation}
    H_{i} = -\sum_{j=0}^{i}a_{i,j}\log a_{i,j} \label{equa:entp}
\end{equation}

Information entropy quantifies the concentration or dispersion of the attention score distribution. A higher entropy value indicates a smoother, more uniform distribution. We will analyze the variation of $H_i$ as the number
of rows i increases. Replacing $Q_i^{T}K_j$ with $w_{i,j}$ for simplification and  substituting Eq. (\ref{equa:ascore}) into Eq. (\ref{equa:entp}), we have
\begin{equation}
H_i = \log \sum_{j=0}^{i} e^{w_{i,j}} - \frac{\sum_{j=0}^{i-1} e^{w_{i,j}} \cdot w_{i,j}}{\sum_{j=0}^{i-1} e^{w_{i,j}}}
\label{equa:pf1 }
\end{equation}
We further assume that $w_{i,j}$ follows the Gaussian distribution. Then, we can deduct that the expectation of $H_{i}$ is
\begin{equation}
    E[{H}_i]  = \log i - \frac{d}{2}.   \label{equa:pf8} 
\end{equation}
Details of the deduction can be found in the appendix \ref{appx:pf1}.
Eq. (\ref{equa:pf8}) shows that expectation of $H_i$ increases with $i$. This indicates that as the number of tokens grows, the attention score distribution becomes flatter, and the maximum attention score decreases. Consequently, tokens in later positions receive lower attention scores overall. To tackle this, we introduce a scaling parameter $\lambda$ to adaptively adjust the smoothness of the distribution according to the number of tokens.
Mathematically, we transform the softmax function from $\text{softmax}(x_i)=\frac{x_i}{\sum_{i=0}^{n}e^{x_i}}$ to step gain sofmax
\begin{equation}
\text{SG-softmax}(x_i,\lambda_{i})=\frac{\lambda x_i}{\sum_{i=0}^{n}e^{\lambda x_i}}.
\end{equation}
Then, the expectation of $H_{i}$ changes into
\begin{equation}
E[{H}_i]  = \log i - \frac{\lambda^{2}d}{2}.   \label{equa:lambda} 
\end{equation}
Assume that the token number of the budget is $k$. The ideal case for a good eviction score is that the there are only $k$ attention score larger than zero. We aim to make expected information entropy equals to the maximum information entropy regardless of the number of total tokens. Therefore, we have
\begin{equation}
\log i - \frac{\lambda^{2}d}{2}=-\sum_{j=0}^{k}\frac{1}{k}\log \frac{1}{k}=-\log \frac{1}{k}. 
\end{equation}
We can figure out that, when the number of total tokens is $i$,
\begin{equation}
    \lambda = \sqrt{\frac{2\log (i/k)}{d}}. \label{equa:pf9} 
\end{equation}
Eq. (\ref{equa:pf9}) gives an estimation for the scaling parameter $\lambda$ in practice.

\subsection{Enhance the Eviction Score via Value-prior}
\label{sec:4.3}
Through the steps outlined above, we have derived unbiased eviction scores. However, since both attention scores and value vectors play a critical role in the computation of self-attention, it is essential to account for their importance. In order to utilize the holistic attention information, we enhance the accumulated attention scores by incorporating the magnitude of the value vectors.

Let $V \in \mathbb{R}^{n \times d}$ denote the value matrix. The square of the $L_2$ norm of the $i^{th}$ token is computed $\nu_i = \|V_i\|^{2}$. However, the value modulus lengths cannot be used directly for enhancement because the values are more independent from each other, which leads to a more discrete distribution of modulus lengths, which is more destructive to the continuity of the text. Therefore, we use mean filtering to optimise the mode length of value to get the enhancement coefficient gamma: 
\begin{equation}
    \gamma = Avgpool(\nu)
\end{equation}
We then normalize it with the maximum value among all $\gamma_{i}$.
\begin{equation}
    \bar{\gamma}_i = \frac{\gamma_{i}}{\max(\gamma)}. \label{equa:pf14.5}
\end{equation}
Unlike the accumulated attention score which measures token importance based on inter-token correlations, 
$\bar{\gamma}_i$ captures the importance derived from the model's parameterized weighting of tokens.  We use it as a prior weight to refine adaptive accumulated attention score: 
\begin{equation}
    \hat{S}_i = \bar{\gamma}_i \cdot S_i \label{equa:pf15}
\end{equation}
The details of the AhaKV algorithm are shown in Algorithm \ref{alg:alg1}.

% {\small
% \begin{algorithm}[H]
% \caption{AhaKV Algorithm}\label{alg:alg1}
% \begin{algorithmic}[1]
%     \STATE Total Cache Budget B($B=B_r+B_s$), Recent Cache Budget $B_r$, Selected Cache Budget $B_s$, Temperature $\beta$
% \end{algorithmic}
% \label{alg1}
% \end{algorithm}
% }
\begin{algorithm}
\small
\caption{AhaKV Algorithm} \label{alg:alg1}
\begin{algorithmic}[1]
    \State Total Cache Budget B($B=B_r+B_s$), Recent Cache Budget $B_r$, Selected Cache Budget $B_s$
    \Function{Prefill}{$Prompt$}
        \For{Layer-i in LLMs}
            \For{Head-m in Layers}
                \State ${Q}_{m,i}, {K}_{m,i}, {V}_{m,i} \in \mathbb{R}^{n\times d}$ 
                \State $A_{m,i} \gets \frac{{Q}_{m,i}\cdot{K}_{m,i}^{T}}{\sqrt{d}}$
                \State $S_{m,i}^{r} \gets$ Recent \hspace{0.1em} $B_r$ \hspace{0.1em} Token
                \State $F_{s} \gets \sum_{Token\in S_{m,i}^{r}} SGsoftmax(A)$
                \State $\gamma \gets Avgpool(\|V\|^{2})$
                \State $\hat{F}_{s} \gets \frac{\gamma}{max(\gamma)}\cdot F_{s}$ 
                \State $S_{m,i}^{s} \gets TopK(F_s, B_s)$
                \State $S_{m,i}^{Prefill} \gets S_{m,i}^{s} \bigcup S_{m,i}^{r}$
            \EndFor
        \EndFor
    \EndFunction
    \Function{Generation}{$S_{m,i}^{Prefill},MaxLength$}
        \State $S^0 \gets S^{prefill}$
        \State $Z_0 \gets$ First\hspace{0.1em} Token
        \State $F_0 \gets \hat{F}_s$
        \For{$t<MaxLength$}
            \For{Layer-i in LLMs}
                \For{Head-m in Layers}
                    \State $K_{m,i}^{t-1},V_{m,i}^{t-1}\gets K_{S_{m,i}^{t-1}},V_{S_{m,i}^{t-1}}$
                    \State $K_{m,i}^{t}\gets(K_{m,i}^{t-1},W_{m,i}^{K}Z_0)$
                    \State $V_{m,i}^{t}\gets(V_{m,i}^{t-1},W_{m,i}^{V}Z_0)$
                    \State $A_{m,i} \gets \frac{{W_{m,i}^{Q}Z_0}_{m,i}\cdot{K}_{m,i}^{T}}{\sqrt{d}}$
                    \State $F_t \gets SGsoftmax(A)+F_{t-1}$
                    \State $S_{m,i}^{r} \gets$ Recent \hspace{0.1em} $B_r$ \hspace{0.1em} Token
                    \State $S_{m,i}^{s} \gets TopK(F_t, B_s)$
                    \State $S_{m,i}^{Generation} \gets S_{m,i}^{s} \bigcup S_{m,i}^{r}$
                \EndFor
            \EndFor
        \EndFor
    \EndFunction
\end{algorithmic}
\end{algorithm}

\section{Experiments}

% \subsection{Setup and Baseline}
% %加引用
% Our experiment are conducted based on 3 widely used basic models, LLaMA \cite{llama}, Qwen \cite{qwen} and Gemma\cite{Gemma}. In order to demonstrate the capability of AhaKV under long text, we chose the LongBench framework \cite{bai2023longbench} for testing. There are 21 datasets in six areas in the LongBench framework, including math, code completion, few-shot learning, multi-document QA, single-document QA, summarization and synthetic tasks. By testing 21 datasets within the LongBench framework, we can effectively demonstrate the efficacy of the AhaKV method. In addition to testing within the LongBench framework, we selected the ARC-E \cite{2018ARC-E}, OpenBookQA \cite{2018OpenBook}, WiC \cite{2019WiC}, and WinoGrande \cite{2020WiC} datasets, demonstrating that AhaKV is effective across a diverse range of datasets. All experiments are conducted in NVIDIA A800 80GB GPU.

% We select H2O as the baseline and primary improvement method. H2O calculates the accumulated attention score and discards the lower value. In addition, we compare AhaKV with several popular KV cache eviction strategies, including Sink \cite{xiao2023efficient}, SnapKV \cite{li2024snapkv}, NACL \cite{chen2024nacl} and TOVA \cite{oren2024transformers}.

\begin{table*}
\centering
\small
% \vspace{-5mm}
\caption{Results in Longbench Datasets}\label{LongBench Experiment}
\label{Comparasion expriment}
\scalebox{0.7}{ \begin{tabular}{l c|c c c c c c c}
\hline
\multicolumn{2}{c|}{Method} & Code Complete & Few Shot & Single-doc QA & Multi-doc QA & Summarization & Passage Retrieval & Average\\
\hline
\multirow{7}{*}{\rotatebox{90}{LLaMA3-8B-Inst}}
&Full Cache&54.47&57.75&41.55&32.33&20.14&54.83&41.94\\
&Sink\cite{xiao2023efficient}&56.59&53.03&30.31&26.88&17.75&26.50&33.55\\
&$H_2O$\cite{zhang2024h2o}&\textbf{58.88}&54.57&38.61&30.76&\textbf{19.75}&41.70&38.93\\
&SnapKV\cite{li2024snapkv}&56.52&56.36&\underline{40.56}&31.06&18.22&54.33&\underline{40.99}\\
&NACL\cite{chen2024nacl}&56.62&54.78&40.32&\underline{31.39}&18.49&\underline{54.33}&40.77\\
&TOVA\cite{oren2024transformers}&53.45&\textbf{57.29}&38.6&30.74&17.63&53.30&40.18\\
&AhaKV&\underline{57.14}&\underline{57.13}&\textbf{41.25}&\textbf{31.85}&\underline{18.66}&\textbf{54.83}&\textbf{41.63}\\
\hline
\multirow{7}{*}{\rotatebox{90}{Qwen2-7B-Inst}}
&Full Cache&59.35&61.60&45.20&34.35&22.65&39.33&42.47\\
&Sink\cite{xiao2023efficient}&56.09&55.41&29.56&26.45&19.43&15.33&32.46\\
&$H_2O$\cite{zhang2024h2o}&52.72&56.14&37.82&31.77&\underline{21.15}&22.67&36.24\\
&SnapKV\cite{li2024snapkv}&\underline{59.08}&\underline{60.37}&\underline{43.99}&\underline{33.33}&20.71&\textbf{38.50}&\underline{41.30}\\
&NACL\cite{chen2024nacl}&58.67&53.67&41.80&32.02&20.83&38.00&39.27\\
&TOVA\cite{oren2024transformers}&56.91&56.65&37.97&30.69&19.00&35.61&37.99\\
&AhaKV&\textbf{59.30}&\textbf{61.04}&\textbf{44.18}&\textbf{33.98}&\textbf{22.02}&\underline{38.33}&\textbf{41.84}\\
\hline
\multirow{7}{*}{\rotatebox{90}{LLAMA2-7B-Chat}}
&Full Cache&55.72&51.99&22.28&18.56&18.55&5.34&27.28\\
&Sink\cite{xiao2023efficient}&53.35&49.26&14.11&15.01&15.66&1.94&23.27\\
&$H_2O$\cite{zhang2024h2o}&42.66&49.32&20.46&16.29&\textbf{18.27}&4.01&24.51\\
&SnapKV\cite{li2024snapkv}&\textbf{55.32}&50.95&\underline{21.50}&17.65&16.82&5.57&\underline{26.39}\\
&NACL\cite{chen2024nacl}&54.50&49.84&19.92&\underline{17.82}&16.97&\textbf{6.54}&26.04\\
&TOVA\cite{oren2024transformers}&53.68&\underline{51.16}&16.10&16.41&15.68&4.34&24.66\\
&AhaKV&\underline{55.06}&\textbf{51.49}&\textbf{22.20}&\textbf{18.16}&\underline{17.12}&\underline{5.57}&\textbf{26.78}\\
\hline
\multirow{7}{*}{\rotatebox{90}{Gemma-7B-Inst}}
&Full Cache&46.91&52.75&33.82&23.81&20.47&27.02&33.25\\
&Sink\cite{xiao2023efficient}&48.84&49.78&20.34&18.63&18.05&15.33&27.18\\
&$H_2O$\cite{zhang2024h2o}&\textbf{50.01}&49.99&29.88&21.29&\textbf{20.13}&22.55&31.09\\
&SnapKV\cite{li2024snapkv}&47.66&\underline{52.45}&\textbf{32.89}&\underline{22.35}&18.84&26.29&\underline{32.39}\\
&NACL\cite{chen2024nacl}&47.09&50.35&31.81&21.93&19.12&\underline{26.69}&31.77\\
&TOVA\cite{oren2024transformers}&45.69&49.20&30.61&21.20&18.78&25.38&30.80\\
&AhaKV&\underline{49.37}&\textbf{53.26}&\underline{32.88}&\textbf{22.93}&\underline{19.38}&\textbf{27.36}&\textbf{33.08}\\
\hline
\end{tabular}
}
% \vspace{-5mm}
\end{table*}

\subsection{Setup and Baseline}
%加引用
Our experiment are conducted based on 3 widely used basic models, LLaMA \cite{llama}, Qwen \cite{qwen} and Gemma\cite{Gemma}. In order to demonstrate the capability of AhaKV under long text, we chose the LongBench framework \cite{bai2023longbench} for testing. There are 21 datasets in six areas in the LongBench framework, including math, code completion, few-shot learning, multi-document QA, single-document QA, summarization and synthetic tasks. By testing 21 datasets within the LongBench framework, we can effectively demonstrate the efficacy of the AhaKV method. In addition to testing within the LongBench framework, we selected the ARC-E \cite{2018ARC-E}, OpenBookQA \cite{2018OpenBook}, WiC \cite{2019WiC}, and WinoGrande \cite{2020WiC} datasets, demonstrating that AhaKV is effective across a diverse range of datasets. Results were averaged over multiple seed results. All experiments are conducted with NVIDIA A800 80GB GPU.

We select H2O\cite{zhang2024h2o} as the baseline and primary improvement method. H2O use accumulated attention score and discards the lower value. In addition, we compare AhaKV with several popular eviction strategies, including Sink \cite{xiao2023efficient}, SnapKV \cite{li2024snapkv}, NACL \cite{chen2024nacl} and TOVA \cite{oren2024transformers}.

\begin{figure*}[t]
\centering
\includegraphics[width=\textwidth]{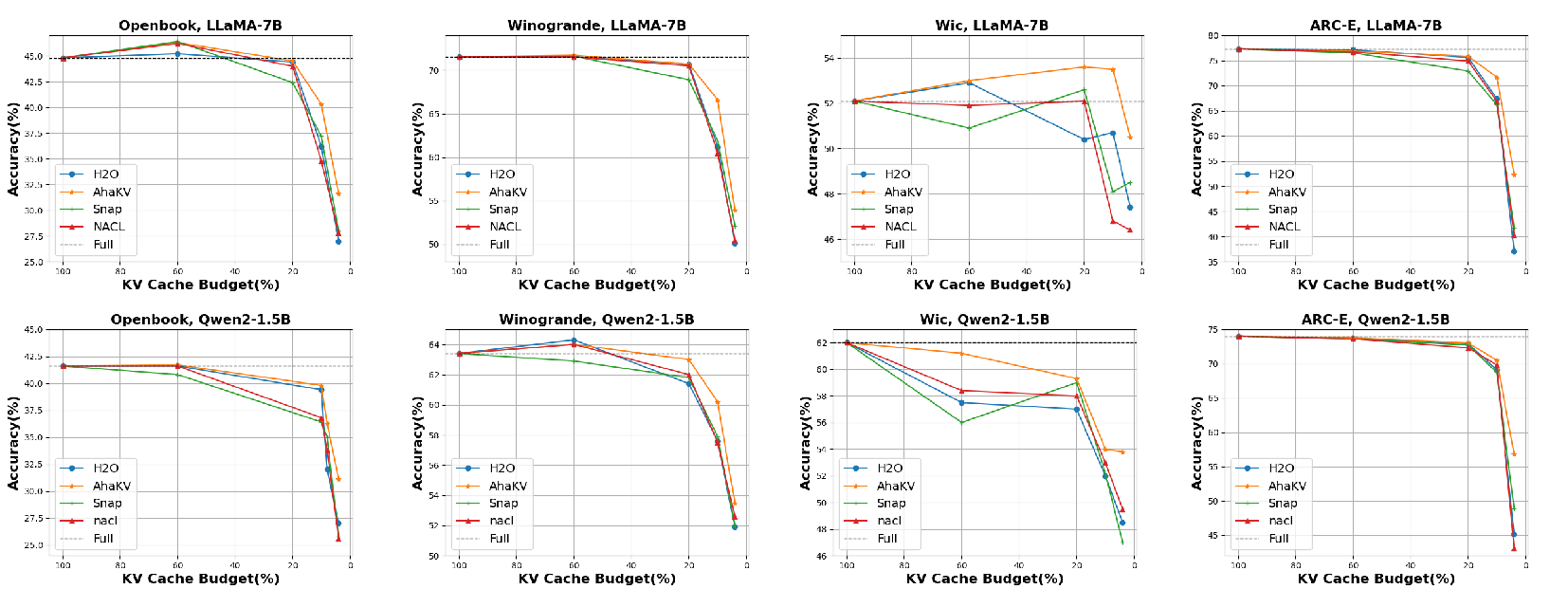}
\caption{Short-text results in different KV budget.}\label{Fig.4}
\end {figure*}
\subsection{Comparison Results}
\subsubsection{Longbench Results}
We conducted extensive testing on the LongBench dataset, which consists of six categories of tasks as in Table \ref{LongBench Experiment}. LLAMA2-7B was assigned a total budget of 720 due to its limited sequence length, while other models received 1000, with a fixed recent budget of 32 applied uniformly across all models. We use Eq. \ref{equa:pf9} to initialize the parameter $\lambda$. As shown in Table \ref{Comparasion expriment}, AhaKV achieved state-of-the-art performance, attaining the highest average accuracy across all tasks.

The AhaKV method demonstrates significant performance advantages across multiple benchmarks. In code-completion tasks, it achieved superior accuracy over the full cache baseline by $2.67\%$ and $2.46\%$ on LLAMA3-8B-Inst and Gemma-7B-Inst, respectively. For few-shot learning scenarios, AhaKV outperformed all competitors on Gemma-7B-Inst with a $0.81\%$ lead over the second-best method, though trailing slightly on LLaMA3-8B-Inst. It consistently dominated both single-document and multi-document QA tasks across all tested models. The method enhances summarization through context-aware extraction of highly relevant tokens and shows strong retrieval capabilities, outperforming rivals by $0.5\%$ and $0.67\%$ on LLaMA3-8B-Inst and Gemma-7B-Inst in passage retrieval. These key metrics demonstrate AhaKV's consistent performance advantages across diverse tasks and model architectures.

% \begin{figure*}[t]
% \centering
% \includegraphics[width=\textwidth]{figures/shortxt.png}
% \caption{Short-text results in different KV budget.}\label{Fig.4}
% \end {figure*}

\subsubsection{Short-text Results}
To validate the effectiveness of AhaKV in processing short texts, we selected the OpenBook, ARC-E, WiC, and WinoGrande datasets for testing and comparison against H2O\cite{zhang2024h2o} and other methods in Figure \ref{Fig.4}.

AhaKV has been evaluated using LLAMA2-7B and Qwen2-1.5B, with results obtained for a KV Budget of 60\%-4\%. The experiments demonstrate that AhaKV can achieve exceptional performance with short text datasets. Notably, as the compression rate increases (resulting in a decreased Budget), AhaKV shows a tendency to deliver higher accuracy compared to H2O and others.

Compared to other methods, SampkeKV consistently outperforms other expulsion schemes in terms of accuracy in the most extreme compressed case (only $4\%$ budget). It shows that AhaKV is able to give a high enough weight to crucial tokens in the text to make them less likely to be evicted.

\begin{table}[h]
\centering
\caption{Ablation study in subdatasets in LongBench of llama2-7B.}
\label{Ablation expriment}
\scalebox{0.65}{ % 将表格宽度调整为单栏宽度
\begin{tabular}{c|c c c c c c}
\hline
Datasets & 2Wiki & TriQA & Multi-EN & lcc & Samsum & Avg \\
\hline
w/o RA&30.26&84.47&36.44&55.77&39.7&49.33\\
w/o SGS&30.5&82.85&37.89&58.0&39.6&49.76\\
w/o VPE&29.54&82.94&33.96&58.16&41.06&49.13\\
AhaKV&30.91&84.34&39.21&58.4&40.1&50.59\\
\hline

\hline
\end{tabular}
}
\end{table}

\subsection{Ablation Study}
In AhaKV, we introduce three ways to improve eviction strategies: \textbf{Recent Accumulation}, \textbf{Step Gain Softmax} and \textbf{Value-Prior Enhance}. To evaluate the orthogonality of the three components and assess their individual impact, we conducted ablation experiments on LLAMA2-7B-Chat, examining each part separately. To simplify the presentation, we use the abbreviations RA for Recent Accumulation, SGS for Step Gain Softmax, and VPE for Value-Prior Enhance in the table.

Table \ref{Ablation expriment} presents ablation results for LLAMA2-7B-Chat on LongBench subsets (one dataset per category). The principal findings are outlined as follows. (i) Removing RA caused a $1.26\%$ average accuracy drop, confirming the necessity of stable cumulative terms to reduce bias. (ii) Eliminating SGS reduced accuracy by $0.83\%$, demonstrating its critical role in eviction strategy enhancement. (iii) Disabling VPE decreased accuracy by $1.46\%$, proving its effectiveness in refining eviction scores to prioritize essential tokens.
% \begin{table}[h]
% \centering
% \caption{Ablation study in subdatasets in LongBench of llama2-7B.}
% \label{Ablation expriment}
% \scalebox{0.65}{ % 将表格宽度调整为单栏宽度
% \begin{tabular}{c|c c c c c c}
% \hline
% Datasets & 2Wiki & TriQA & Multi-EN & lcc & Samsum & Avg \\
% \hline
% w/o RA&28.7&84.17&31.01&57.38&39.49&48.15\\
% w/o SGS&29.78&81.23&29.67&50.32&40.37&46.25\\
% w/o VPR&29.54&82.94&33.96&58.16&41.06&49.13\\
% AhaKV&29.96&83.61&34.6&58.4&40.55&49.42\\
% \hline

% \hline
% \end{tabular}
% }
% \end{table}

% \begin{figure}[h]
% \centering
% \includegraphics[width=5cm]{latex/figures/Temperature.png}
% \caption{Temperature.}\label{Fig.4}
% \end {figure}

\section{Conclusion}
In this paper, we address the problem of KV cache compression by proposing AhaKV, a KV cache eviction algorithm that can identify important tokens and reason with a fixed KV cache budget, solving the problem of KV cache growth at inference. Our approach selects key tokens based on the attention score, and proposes Step Gain Softmax and Sample Attention to eliminate the attention bias problem caused by causal mask. Experimental evaluations show that AhaKV achieves good performance on both short text and long text datasets.

\section*{Limitations}

The main limitations of the approach presented in the paper are the following: firstly, due to resource constraints, we did not conduct our experiments on longer texts. However, we have derived our method and evaluated our model on datasets of different lengths in the paper, and we believe that AhaKV can be adapted to reasoning with longer texts. Second, for the choice of a constant number of accumulators in prefill, we choose the recent rows in the attention score for accumulation; in practice, we do not think this choice is unique. For example, we could also choose the Topk strategy to select the same number of cumulants.

\section*{Ethics Statement}
In this study, we utilize open-source data and technologies, which significantly mitigate privacy concerns. With the polish of the AI assistant, we more clearly define the problem and illustrate the approach. Our novel approach focuses on enhancing the understanding of model contexts and improving inference efficiency, with the goal of creating accessible and highly effective models capable of handling extended contexts. This strategy is expected to advance the openness of NLP technologies and facilitate their practical deployment in diverse applications. Crucially, our methodology is independent of the training process, ensuring that it does not perpetuate or introduce biases into the models. By emphasizing state-of-the-art, resource-efficient techniques, our research contributes to the development of a more open and automated AI, pushing the boundaries of artificial intelligence while ensuring that the benefits of these advancements are widely accessible and applicable across various fields. This represents a significant step toward a more inclusive and automated AI-driven future.

\bibliography{custom}

\appendix
{
\section{Derivation of the mathematical expectation of $H_i$} \label{appx:pf1}

The mathematical expectation of $H_i$ is shown as Eq.\ref{equa:apf4}
\begin{align}
    logi + logE[e^{w_{i,j}}] - \frac{E[e^{w_{i,j}}\cdot w_{i,j}]}{E[e^{w_{i,j}}]}
    \label{equa:apf4}
\end{align}

Let $x=w_{i,j}$ and assume that x is a Gaussian distribution with mean $\mu$ and variance $\sigma^2$. Symbolically, $x \sim N(\mu, \sigma^2)$. Its probability density function is shown as Eq. \ref{equa:ap1}.
\begin{equation}
    f(x) = \frac{1}{\sigma \sqrt{2 \pi}}e^{- \frac{(x-\mu)^2}{2 \sigma^2}} \label{equa:ap1}
\end{equation}

The mathematical expectation of $e^x$ in Gaussian distribution is shown as the Eq. \ref{equa:ap2}.
\begin{equation}
    E(e^x) = \frac{1}{\sigma \sqrt{2 \pi}}\int_{-\infty}^{\infty}e^x\cdot e^{- \frac{(x-\mu)^2}{2 \sigma^2}}dx \label{equa:ap2}
\end{equation}

Combine the Equation \ref{equa:ap2} to the Equation \ref{equa:ap3}.
\begin{equation}
    E(e^x) = \frac{1}{\sigma \sqrt{2 \pi}}\int_{-\infty}^{\infty}e^{x - \frac{(x-\mu)^2}{2 \sigma^2}}dx \label{equa:ap3}
\end{equation}

To simplify the Equation \ref{equa:ap3}, we adjust the $x - \frac{(x-\mu)^2}{2 \sigma^2}$ as follow.
\begin{equation}
    \mu + \frac{\sigma^2}{2} - \frac{[(x-\sigma^2)-\mu]^2}{2\sigma^2} \label{equa:ap5}
\end{equation}

Combine the Equations \ref{equa:ap3} and \ref{equa:ap5} to Equation \ref{equa:ap7}.
{\begin{equation}
    e^{\mu+\frac{\sigma^2}{2}}\int_{-\infty}^{\infty}\frac{1}{\sigma \sqrt{2 \pi}}e^{- \frac{[(x-\sigma^2)-\mu]^2}{2\sigma^2}}dx \label{equa:ap7}
\end{equation}}

Denote $t$ as $x-\sigma^2$, the Equation \ref{equa:ap7} could be transformed as the Equation \ref{equa:ap8}.
\begin{equation}
    e^{\mu+\frac{\sigma^2}{2}}\int_{-\infty}^{\infty}\frac{1}{\sigma \sqrt{2 \pi}}e^{- \frac{(t-\mu)^2}{2\sigma^2}}dt \label{equa:ap8}
\end{equation}

The $\int_{-\infty}^{\infty}\frac{1}{\sigma \sqrt{2 \pi}}e^{- \frac{(t-\mu)^2}{2\sigma^2}}dx$ is the cumulative distribution function of random variable which obey the Gaussian distribution with mean $\mu$ and variance $\sigma$. The result of the integral is 1. Thus, the mathematical expectation of $e^x$ is represented as follow.
\begin{equation}
    E(e^x) = e^{\mu+\frac{\sigma^2}{2}}
\end{equation}

The computation of $E(xe^x)$ is shown as the Equation \ref{equa:ap9}.
\begin{align}
    E(xe^x) &= \frac{1}{\sigma \sqrt{2 \pi}}\int_{-\infty}^{\infty}xe^x\cdot e^{- \frac{(x-\mu)^2}{2 \sigma^2}}dx \label{equa:ap9} \\
    &= \frac{1}{\sigma \sqrt{2 \pi}}\int_{-\infty}^{\infty}x\cdot e^{x- \frac{(x-\mu)^2}{2 \sigma^2}}dx \label{equa:ap10}
\end{align}

To simplify the Equation \ref{equa:ap10}, we adjust the $x - \frac{(x-\mu)^2}{2 \sigma^2}$ as shown in the Equation \ref{equa:ap5}. Thus, the Euqation \ref{equa:ap10} is transformed into the Equation \ref{equa:ap11}.
\begin{equation}
    e^{\mu+\frac{\sigma^2}{2}}\int_{-\infty}^{\infty}\frac{x}{\sigma \sqrt{2 \pi}}e^{- \frac{[(x-\sigma^2)-\mu]^2}{2\sigma^2}}dx \label{equa:ap11}
\end{equation}

Denote $u$ as $x-\sigma^2-\mu$, the Equation \ref{equa:ap11} is transformed into the Equation \ref{equa:ap13}.
\begin{equation}
    \frac{e^{\mu+\frac{\sigma^2}{2}}}{\sigma \sqrt{2 \pi}}[\int_{-\infty}^{\infty}(\sigma^2+\mu) e^{- \frac{u^2}{2\sigma^2}}du + \int_{-\infty}^{\infty}u e^{- \frac{u^2}{2\sigma^2}}du] \label{equa:ap13}
\end{equation}

The first term of the Equation \ref{equa:ap13} is similar to the integral of the Gaussian distribution and its result is $\sigma \sqrt{2\pi}$. The second term of the Equation \ref{equa:ap13} is an odd function and its integral is 0. In summary, the result of the Equation \ref{equa:ap13} is shown as follow.
\begin{equation}
    E(xe^x) = e^{\frac{\sigma^2}{2}+\mu}(\mu+\sigma^2)
\end{equation}

Combining the above derivations, we can obtain the result of Eq. \ref{appx:pf1} as shown in Eq. \ref{appx:pf21}.
\begin{equation}
    \log i - \frac{d}{2} \label{appx:pf21}
\end{equation}

\begin{figure}
    \centering
    \includegraphics[width=\linewidth]{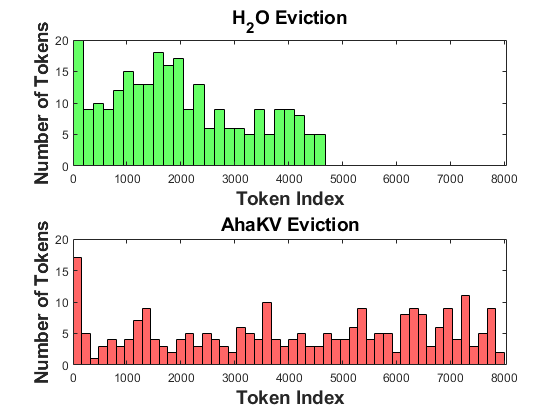}
    \caption{Retain token index in Qwen2-7B-Inst.}
    \label{fig:5}
\end{figure}

\section{Additional experiments on different scale models}
In the previous paper we mainly showed the effect of AhaKV on the 7B LLM. We add experiments in Table \ref{Addicition LongBench Experiment} on models at different scales of 1.5B, 4B and 14B to analyse whether our approach generalises to models at different scales. The settings in the experiments are all consistent with those in Table \ref{LongBench Experiment}. 

According to the experiments in Table \ref{Addicition LongBench Experiment} we can see that AhaKV achieves sota results on models of different scales. It shows that our method can be generalised to models of any scale.

\begin{table*}
\centering
\small
% \vspace{-5mm}
\caption{Results in Longbench Datasets on Different Scale Models}\label{Addicition LongBench Experiment}
\label{Comparasion expriment}
\scalebox{0.9}{ \begin{tabular}{l c|c c c c c c c}
\hline
\multicolumn{2}{c|}{Method} & Code Complete & Few Shot & Single-doc QA & Multi-doc QA & Summarization & Passage Retrieval & Average\\
\hline
\multirow{7}{*}{\rotatebox{90}{Qwen2-1.5B-Inst}}
&Full Cache&39.93&54.77&36.53&28.3&22.36&8.33&32.03\\
&Sink&39.26&49.5&22.64&20.52&17.45&5.33&25.47\\
&$H_2O$&38.9&49.09&23.52&21.34&18.07&5.17&25.78\\
&SnapKV&40.31&54.59&33.81&26.15&18.73&8.16&30.39\\
&NACL&39.59&51.91&31.48&25.46&18.87&8.33&29.29\\
&TOVA&37.61&49.91&29.14&23.83&17.18&6.00&27.31\\
&AhaKV&39.99&54.68&34.60&29.10&19.81&8.83&31.39\\
\hline
\multirow{7}{*}{\rotatebox{90}{Qwen1.5-4B-Chat}}
&Full Cache&34.29&57.13&40.98&31.66&22.80&11.83&34.02\\
&Sink&34.24&52.73&26.29&21.76&18.51&7.66&27.08\\
&$H_2O$&36.61&51.48&37.81&29.06&21.52&10.33&31.61\\
&SnapKV&35.10&55.69&40.09&30.48&20.13&10.83&32.77\\
&NACL&35.39&53.18&39.05&29.14&19.93&11.17&31.89\\
&TOVA&33.1&51.56&37.4&28.54&19.65&10.00&30.70\\
&AhaKV&35.30&55.37&40.25&31.10&20.78&11.50&33.09\\
\hline
\multirow{7}{*}{\rotatebox{90}{Qwen1.5-14B-Chat}}
&Full Cache&56.64&60.24&42.65&37.85&22.24&53.66&44.11\\
&Sink&55.10&54.39&26.80&26.94&19.03&16.00&31.75\\
&$H_2O$&56.07&57.37&39.84&36.04&22.83&51.91&42.48\\
&SnapKV&56.97&59.22&41.50&36.54&21.04&53.33&43.20\\
&NACL&56.67&57.15&40.67&36.15&21.36&53.5&42.66\\
&TOVA&52.63&57.15&39.02&35.06&20.8&52.33&41.24\\
&AhaKV&61.79&59.45&41.22&36.88&21.59&53.67&43.87\\
\hline
\end{tabular}
}
\end{table*}

\section{Additional experiments on retaining token comparisons} 
\label{appx:C}
In Sec \ref{sec:3.1}, we show a comparison of token retention for H2O and AhaKV in the LLaMA2-7B-Chat model. In fact, this comparison is so extensive that we added the LLaMA3-8B-Inst and Qwen2-7B-Inst comparisons shown in Figures \ref{fig:5} and \ref{fig:6}. As can be seen from the figures \ref{fig:5} and \ref{fig:6}, the bias of the H2O-based approach very clearly favours the retention of the tokens located on the left side of the sequence and leads to the neglect of the tokens located on the right side of the sequence, which can indicate that our proposed position bias is factually present.

% \begin{figure}
%     \centering
%     \includegraphics[width=\linewidth]{latex/figures/fig5.png}
%     \caption{Retain token index in Qwen2-7B-Inst.}
%     \label{fig:5}
% \end{figure}

\begin{figure}
    \centering
    \includegraphics[width=\linewidth]{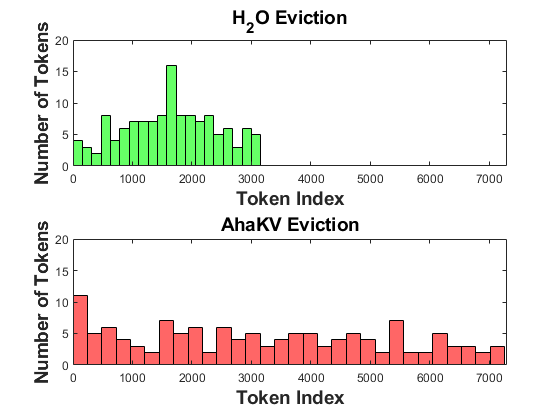}
    \caption{Retain token index in LLaMA3-8B-Inst.}
    \label{fig:6}
\end{figure}

\section{Why do we need to remove bias from evictions?}
We fully justify the reasons for the bias of $H_2O$-base's eviction method with mathematical theory in the main text, and how we can perform unbiased eviction of long texts with AhaKV. To explain how reducing the bias in eviction affects the input to LLM, we show the distribution of paragraph indexes for the retrieval task. 

\begin{figure}
    \centering
    \includegraphics[width=\linewidth]{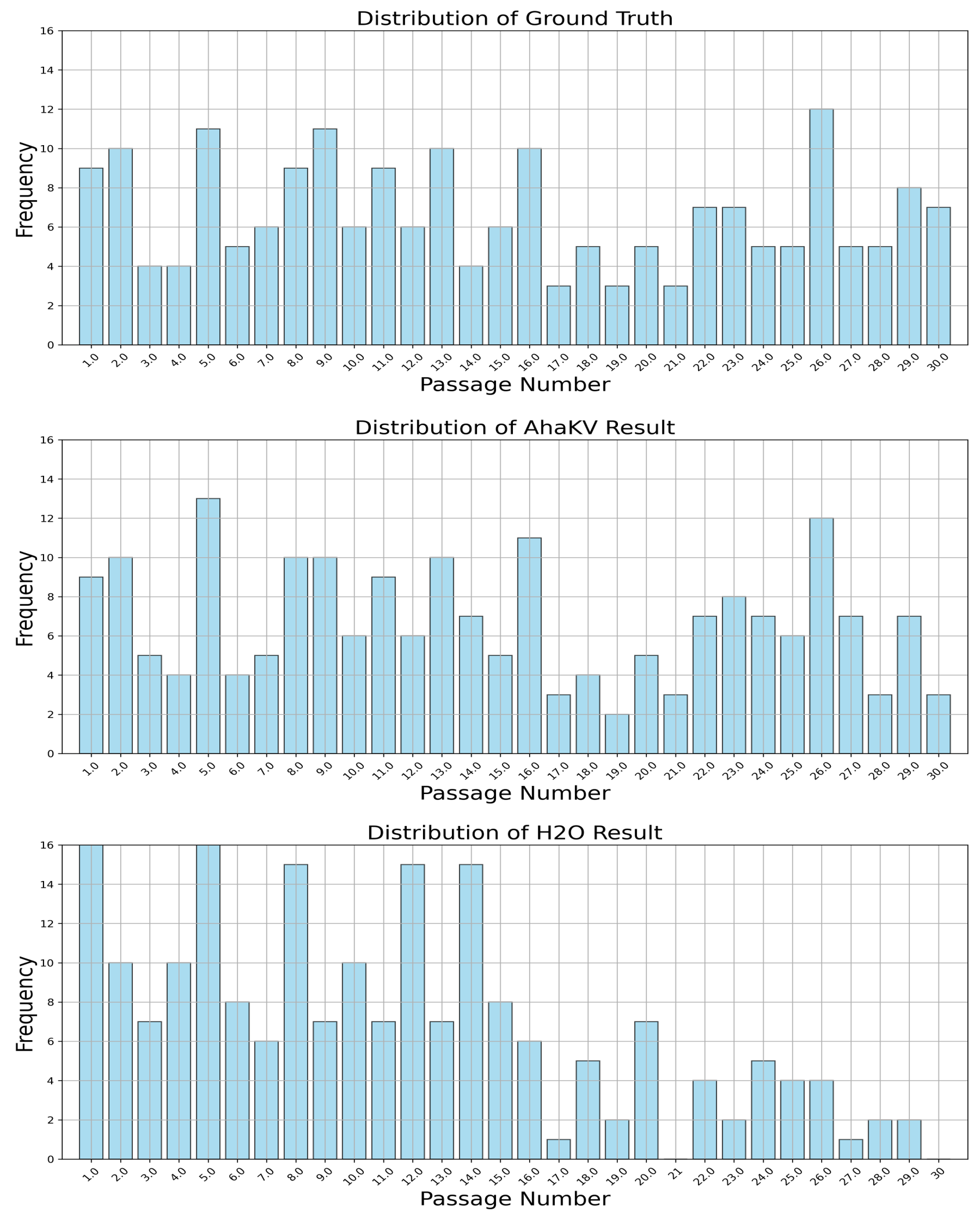}
    \caption{Distribution of the results of different eviction strategies in the passage retrieval task.}
    \label{fig:appx1_pr}
\end{figure}

The retrieval task needs to find the most relevant documents to the query based on the content of the query. While the KV eviction strategy is to find the most relevant KV pairs for the generation task, in fact, the goal between the two is highly consistent, so it is more instructive to study the performance of the eviction method in the retrieval task compared to other tasks. In Figure \ref{fig:appx1_pr} we show the distribution of paragraph indexes obtained with the LLaMA3-8B-Inst model for different KV eviction strategies on the passage\_retrieval\_zh task. It can be seen that the distribution of labelled paragraphs in Ground Truth can be roughly seen as a uniform distribution with retrieval ranges present from 1 to 30 paragraphs. The retrieval results of AhaKV show that our approach does a good job of preserving the key information in the global context, and the distribution of the retrieval results is basically in line with the Ground Truth. The retrieval results of H2O results are mainly concentrated in the first half, while the retrieval results of the second half are significantly smaller than the first half. It fully demonstrates the short-sightedness that results from biased eviction strategies. 

\section{The comparison of AhaKV and FlashAttention} \label{appx:pf2}
As a general LLM inference acceleration method, FlashAttention\cite{dao2022flashattention} has been widely applied in many scenarios. AhaKV enables compatibility with FlashAttention, and we will discuss how AhaKV achieves compatibility with it. 

Based on previous implementations of the KV cache eviction method \cite{chen2024nacl,zhang2024h2o,li2024snapkv}, the eviction score is computed additionally and does not interfere with the calculation of self-attention. In the self-attention calculation, we can still use FlashAttention to achieve efficient inference. As for the additional eviction score, due to the recent accumulation, we only need to compute a small number of attention scores of the nearest $B_r$ rows, and the additional computation accounts for a very small fraction of the total LLM inference, which does not significantly affect the inference speed. To provide the speedup effect of our algorithm in combination with FlashAttention, we record the inference time of the AhaKV algorithm for different token budgets on Qwen2-7B-Inst in Figure \ref{fig:7}. The input length used for the experiments is 32K, while the output length is 512. Where full denotes the speed of inference using only FalshAttention, and all the rest of the results are the speed of KV cache eviction using ahakv under FalshAttention.

\begin{figure}
    \centering
    \includegraphics[width=\linewidth]{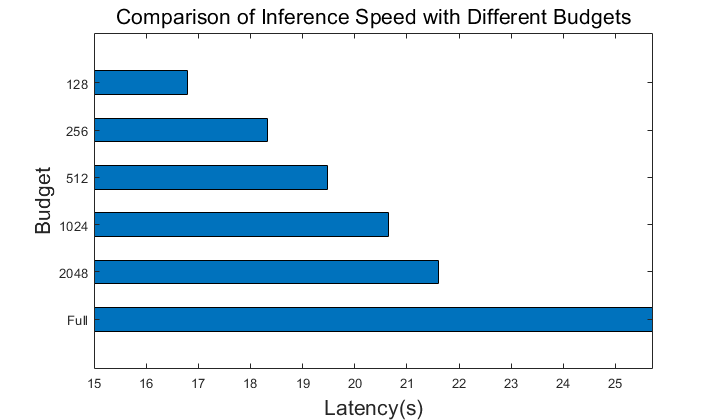}
    \caption{Inference speed in different budget in Qwen2-7B-Inst.}
    \label{fig:7}
\end{figure}

According to Figure \ref{fig:7}, using our method can significantly reduce the inference latency. And it can be seen that the improvement of the inference time becomes more and more obvious as the budget decreases. In the above 32K+512 inference, the KV cache generated using the original FlashAttention takes up 5.98G of memory consumption, while the AhaKV approach of retaining the critical 2048 KV pairs for each header takes up only 0.39G of memory consumption. It undoubtedly reduces the deployment cost of LLM substantially and makes it easier to apply in low-resource device scenarios.

}

% Bibliography entries for the entire Anthology, followed by custom entries
%\bibliography{anthology,custom}
% Custom bibliography entries only
% \bibliography{custom}

\end{document}